\documentclass{spie}

\usepackage{graphicx}
\usepackage[cmex10]{amsmath}
\usepackage{amsfonts} 
\usepackage{amssymb}
\usepackage{stmaryrd}
\usepackage{subfigure}
\usepackage{algorithm,algorithmic}
\usepackage{multirow}
\usepackage{array}

\def\D{{\mathcal{D}}}

\title{Fried deconvolution}

\author{J\'er\^ome~Gilles and~Stanley~Osher
\skiplinehalf
UCLA Department of Mathematics, 520 Portola Plaza, Los Angeles, CA 90095, USA.}

\authorinfo{Further author information: (Send correspondence to J\'er\^ome Gilles)\\
J\'er\^ome Gilles.: jegilles@math.ucla.edu, Telephone: 1 310 794 7737, http://www.math.ucla.edu/$\sim$jegilles\\ Stanley Osher.: sjo@math.ucla.edu, Telephone: 1 310 825 1758}

\begin{document}
\maketitle

\begin{abstract}
In this paper we present a new approach to deblur the effect of atmospheric turbulence in the case of long range imaging. Our method is based on an 
analytical formulation, the Fried kernel, of the atmosphere modulation transfer function (MTF) and a framelet based deconvolution algorithm. An 
important parameter is the refractive index structure which requires specific measurements to be known. Then we propose a method which provides a good 
estimation of this parameter from the input blurred image. The final algorithms are very easy to implement and show very good results on both simulated 
blur and real images.
\end{abstract}

\keywords{blind image deconvolution, Fried kernel, atmospheric blur, parameter estimation}


\section{Introduction}
In long range imaging, the blur due to the atmosphere on the acquired image is non negligible. Two main issues can be observed depending on the level of 
turbulence: a blurring effect and a geometrical distortion effect. From a physics point of view, these effects are clearly correlated but very hard to 
model because of the different physical parameters (temperature, wind, humidity, wavelength, $\ldots$). From the image processing point of view, it is 
easier to consider these two effects as two separate operators which we want to invert. In \cite{Mao2011}, the authors address the problem of correcting 
the geometrical distortions from an input sequence and use some usual deblurring algorithm based on a Gaussian kernel assumption at the end of the 
process to deal with the blur. This final processing about blur does not significantly improve the final image from the geometrically corrected one 
because of the ignorance of the blur kernel.\\

At this moment, two methods are possible. The first one is to use some blind deconvolution algorithm but in the general case this kind of algorithm 
remains a very active field of research and no ``simple'' algorithms are available. Another method is to find some analytic expression which model the 
blurring effect of the atmosphere. Surprisingly, this model exists since $\ldots$ 1966! David Fried, well known in the optical science community for the 
definition of Fried's seeing diameter used to characterize the optical resolution limit, proposed in \cite{Fried1966} an analytical formulation to model 
the Modulation Transfer Function (MTF) of the atmosphere. This work was recently revisited by D.~Tofsted in \cite{Tofsted2011}. It appears that Fried's 
MTF was not well-accepted by the optical science community. But recently, some experiments based on field trials show the effectiveness of this MTF to 
model real phenomena \cite{Buskila2004}.\\

In this paper we propose to use the Fried kernel to deal with the atmospheric blur (the geometrical distortions are not taken into account here; however 
in the experiments we show some results using the output of the algorithm proposed in \cite{Mao2011} as input to the deconvolution). In section 
\ref{sec:FK}, we give the analytical formulation of the Fried kernel. Some details are given about the different parameters and will show that if most 
of them are linked with the optical system, the last one, $C_n^2$, is linked with the level of turbulence of the atmosphere and is difficult to know in 
practice. In section \ref{sec:FD}, we recall the nonblind framelet based deconvolution algorithm of \cite{Cai,Cai2009b} which we will use in the 
following sections. In section \ref{sec:NBFD}, we present a very simple nonblind Fried deconvolution assuming that the $C_n^2$ parameter is known. 
Some properties of the Fried deconvolution will be observed from different experiments done on simulated and real images. In section \ref{sec:BFD}, we 
address the most practical case where the $C_n^2$ parameter is unknown. We propose a criteria to find a good estimate of this parameter. Finally we 
introduce a blind Fried deconvolution algorithm. The different experiments show the effectiveness of the propose method on both simulated and real 
images.

\section{Fried kernel}\label{sec:FK}
Based on papers \cite{Fried1966,Tofsted2011}, the Fried kernel can be viewed as a combination of two terms. One, $M_0$ corresponds to a combination of 
the system plus atmosphere MTFs when the turbulence is negligible. The second, $M_{SA}$, also called the short-term exposure MTF, models the impact (in 
term of blur) of phase-tilt due to the turbulence. Denoting by $\omega$ the spatial frequency (in 2D we consider an isotropic kernel and $\omega$ is the 
frequency modulus), $M_0(\omega)$ can be expressed by
\begin{equation}
M_0(\omega)=
\begin{cases}
\frac{2}{\pi}\left(\arccos(\omega)-\omega\sqrt{1-\omega^2}\right) \qquad \omega<1\\
0 \qquad \omega>1
\end{cases}
\end{equation}
and $M_{SA}(\omega)$ is given by
\begin{equation}
M_{SA}(\omega)=\exp\left\{-(2.1X)^{5/3}(\omega^{5/3}-V(Q,X)\omega^2)\right\}
\end{equation}
If we denote
\begin{itemize}
\item $D$: the system entrance pupil diameter (we recall from the geometrical optics that $D=f/N$ where $f$ is the focal length and $N$ the optics F-number),
\item $L$: the path length (distance from the sensor to the acquired scene),
\item $\lambda$: the wavelength,
\item $C_n^2$: the refractive index structure representing the turbulence level of the atmosphere \cite{Tunick2005},
\end{itemize}

we can define the following quantities: $k=\frac{2\pi}{\lambda}$, the coherence diameter $r_0$ defined by $r_0=2.1\rho_0$ where $\rho_0=1.437(k^2LC_n^2)^{-3/5}$ (the coherence length),  $P=\sqrt{\lambda L}$, $Q=\frac{D}{P}$, $X=\frac{D}{r_0}$. \\

Finally, the quantity $V(Q,X)$ is defined as follows
\begin{equation}
V(Q,X)=A+\frac{B}{10}\exp\left\{-\frac{(x+1)^3}{3.5}\right\}
\end{equation}
where $x=\log_{10}(X)$, $q=\log_2(Q)$ and
\begin{equation}
A=
\begin{cases}
0.840+0.116\Sigma_{qa} \quad \text{with} \quad qa=1.35(q+1.50) \\
\qquad \qquad \qquad \qquad \qquad \qquad \text{if} \; q<-1.50\\
0.840+0.280\Sigma_{qc} \quad \text{with} \quad qc=0.51(q+1.50) \\
\qquad \qquad \qquad \qquad \qquad \qquad \text{if} \; q\geqslant -1.50\\
\end{cases}
\end{equation}
and $\Sigma_q=\frac{e^q-1}{e^q+1}$. The coefficient $B$ is defined by
\begin{equation}
B=0.805+0.265\Sigma_{qb} \quad \text{with} \quad qb=1.45(q-0.15)
\end{equation}

\begin{figure}[!t]
\begin{center}
\begin{tabular}{cc}
  & \multirow{2}{*}{\includegraphics[scale=0.39]{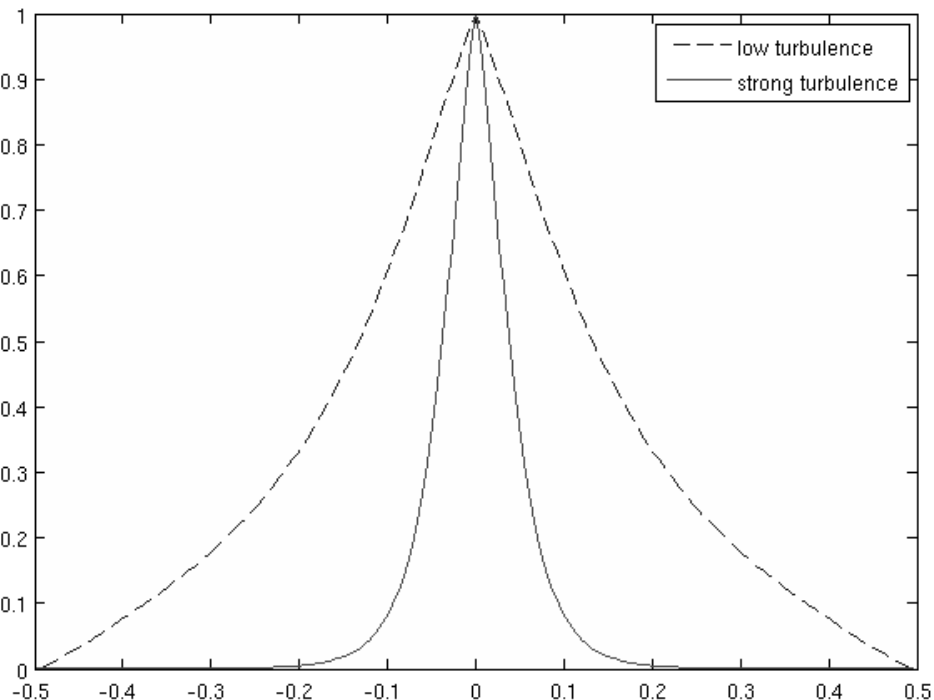}} \\
  \includegraphics[scale=0.22]{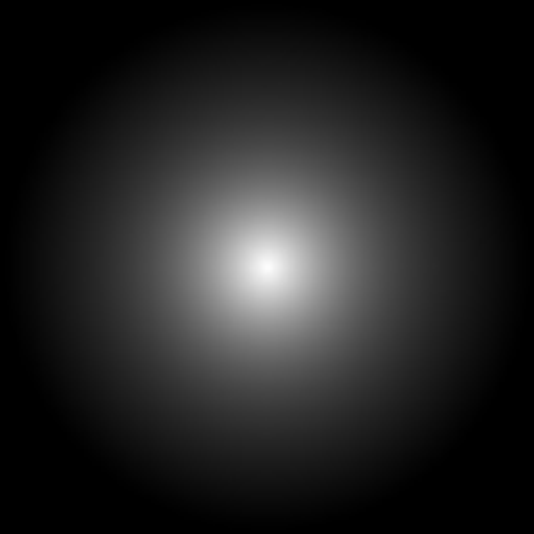} & \\
\includegraphics[scale=0.22]{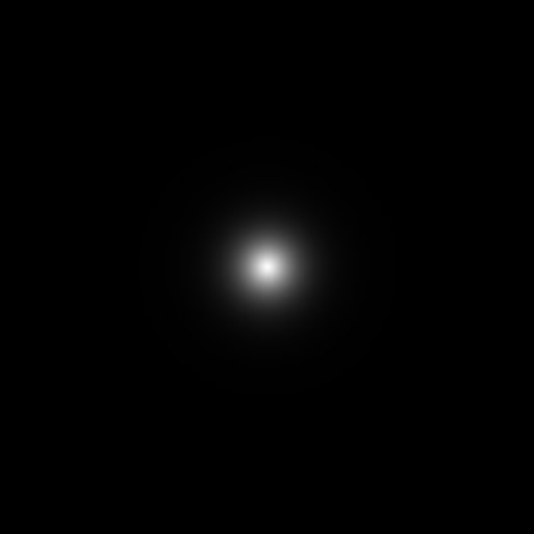} &
\end{tabular}
\end{center}
\caption{Examples of Fried kernel (see text for parameter's values).}
\label{fig:friedkernel}
\end{figure}

\begin{figure}[!t]
\begin{center}
\begin{tabular}{cc}
\includegraphics[scale=0.45]{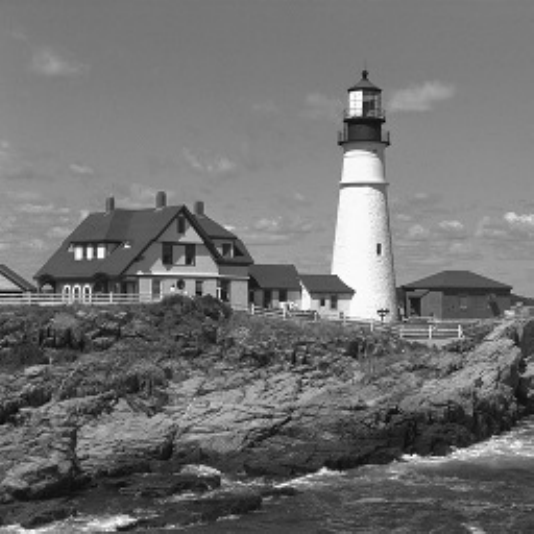} &
\includegraphics[scale=0.45]{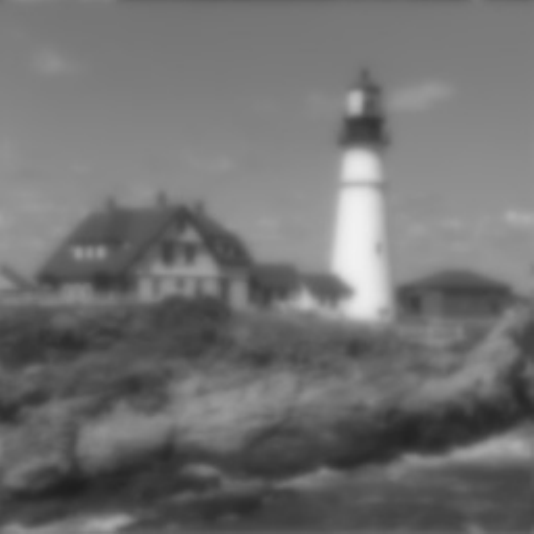}
\end{tabular}
\end{center}
\caption{Examples of an original and its corresponding blurred images.}
\label{fig:ExBlur}
\end{figure}

Finally, in the Fourier domain Fried's MTF $M_F(\omega)$ is the product of $M_0(\omega)$ and $M_{SA}(\omega)$:
\begin{equation}
M_F(\omega)=M_0(\omega)M_{SA}(\omega)
\end{equation}
Practically, this kernel depends only on four parameters: $D,L,\lambda$ and $C_n^2$. The three first clearly depend on the acquisition system and the imaging scene. The last parameter, $C_n^2$ represents the turbulence level and, according to measurement, \cite{Tunick2005}, is generally in the range $[10^{-16}m^{-2/3},10^{-12}m^{-2/3}]$ corresponding respectively to weak and strong turbulence. Two illustrations of the Fried kernel are given with their 1D profile in Fig.~\ref{fig:friedkernel} in the case $D=0.05$m, $L=500$m, $\lambda=700$nm (visible spectra) for $C_n^2=5e^{-14}m^{-2/3}$ and $C_n^2=5e^{-13}m^{-2/3}$. Fig.~\ref{fig:ExBlur} shows an original image in the left and its blurred version on right with parameters $D=0.05$m, $L=500$m, $\lambda=700$nm and $C_n^2=2e^{-13}m^{-2/3}$.


\section{Framelet deconvolution}\label{sec:FD}
Different methods are available in the literature to deal with the nonblind deconvolution problem. The two main categories are algorithms based on a Bayesian framework and algorithms using some functional with regularization constraints. A good starting point about those methods can be found in the book \cite{Campisi2007}. \\

A very efficient deconvolution algorithm when the blur kernel is assumed to be known is the one proposed by Cai et al. in \cite{Cai,Cai2009b}. This algorithm aims to find the image $\hat{u}$ which has a sparse representation in a framelet expansion. We denote $\D$ and $\D^T$ respectively the framelet decomposition and framelet reconstruction operators, see \cite{Cai2009b} for more details. Following the framelet properties we have $\D^T\D=I$ (tight frame) where $I$ stands for the identity. Denoting $g$ the acquired blurred image, $A$ the known blur kernel, the nonblind deconvolution is done by finding $\hat{u}$ which minimizes the following functional

\begin{equation}
\hat{u}=\arg\min \|\D u\|_1+\frac{\mu}{2}\|Au-g\|_2^2
\end{equation}

To solve this minimization problem, we set $d=\D u$ (the framelet expansion of $u$) and we use the split Bregman iteration (see \cite{Goldstein2009} for details):
\begin{equation}
\begin{cases}
u^{k+1}=\arg\min \frac{\mu}{2}\|Au-g\|_2^2+\frac{\eta}{2}\|d^k-\D u-b^k\|_2^2\\
d^{k+1}=\arg\min \|d\|_1+\frac{\eta}{2}\|d-\D u^{k+1}-b^k\|_2^2\\
b^{k+1}=b^k+\D u^{k+1}-d^{k+1}
\end{cases}
\end{equation}
It as been shown that solving for $d^{k+1}$ is equivalent to use the shrinkage operator:
\begin{align}
d^{k+1}&=shrink(\D u^{k+1}+b^k,1/\eta)\\
&=sign(u)\max(0,|u|-1/\eta)
\end{align}
Solving for $u$ is a classic $L^2$ minimization problem which gives
\begin{equation}
(\mu A^*A+\eta I)u-\mu A^*g-\eta \D^T(d^k-b^k)=0
\end{equation}
And we use the Fourier transform to solve this problem:
\begin{equation}\label{eq:nbfdec}
\hat{U}^{k+1}=\left(\mu |\hat{A}|^2+\eta\right)^{-1}(\mu \bar{\hat{A}}\hat{G}+\eta \widehat{\D^T(d^k-b^k)})
\end{equation}
Where the hat symbol stands for the direct Fourier transforms and multiplications are considered pointwise. The corresponding method is summarized in algorithm \ref{algo:nbframan}.
\begin{algorithm}
\caption{Nonblind Frame deconvolution}
\label{algo:nbframan}
\begin{algorithmic}
\STATE $u^0=0,d^0=0,b^0=0,k=0,N_{iter}=$ maximum number of iterations
\WHILE {`'$k<N_{iter}$''}
\STATE Update $u^{k+1}$ by using the inverse Fourier transform of equation (\ref{eq:nbfdec})
\STATE $d^{k+1}=shrink(\D u^{k+1}+b^k,1/\lambda)$
\STATE $b^{k+1}=b^k+\D u^{k+1}-d^{k+1}$
\ENDWHILE
\end{algorithmic}
\end{algorithm}

\section{Nonblind Fried deconvolution}\label{sec:NBFD}
\subsection{Algorithm description}
In this section, we assume that physical parameters $D,L,\lambda,C_n^2$ are known. From the material of the two previous sections, we can easily design a nonblind Fried deconvolution algorithm. First, we build the corresponding Fried kernel and then we use it in place of $A$ in the sparse framelet deconvolution algorithm. The nonblind Fried deconvolution, which is quite simple, is presented in algorithm \ref{algo:nbfried}.

\begin{algorithm}
\caption{Nonblind Fried deconvolution}
\label{algo:nbfried}
\begin{algorithmic}
\STATE - $D,L,\lambda, C_n^2$ are known. Fix the regularization parameters $\mu,\eta$ and the maximum number of iterations $N_{iter}$.
\STATE - Build the Fried kernel $M_F(\omega)$ from these parameters.
\STATE - Use the framelet nonblind deconvolution algorithm to find the restored image $\hat{u}$.
\end{algorithmic}
\end{algorithm}

\subsection{Experiments}
\begin{figure}[!t]
\begin{center}
\begin{tabular}{ccc}
\includegraphics[scale=0.45]{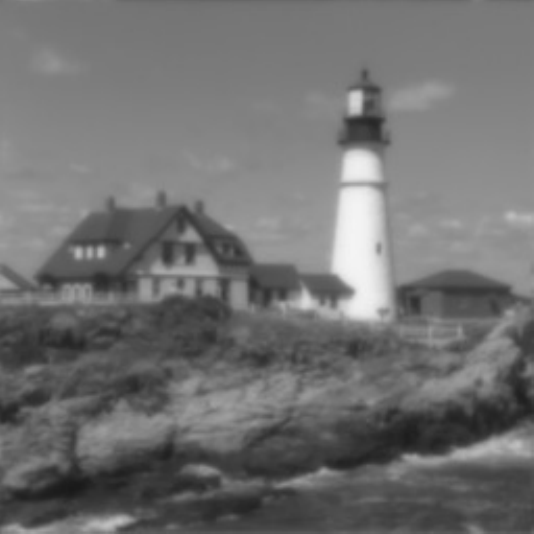} &
\includegraphics[scale=0.45]{Bphare2e13} &
\includegraphics[scale=0.45]{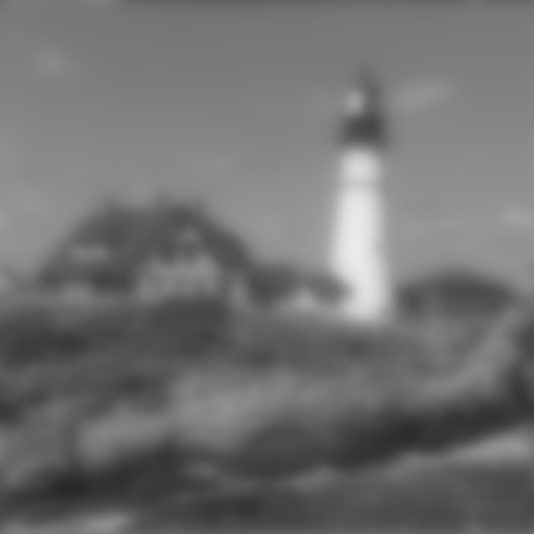} \\
\includegraphics[scale=0.45]{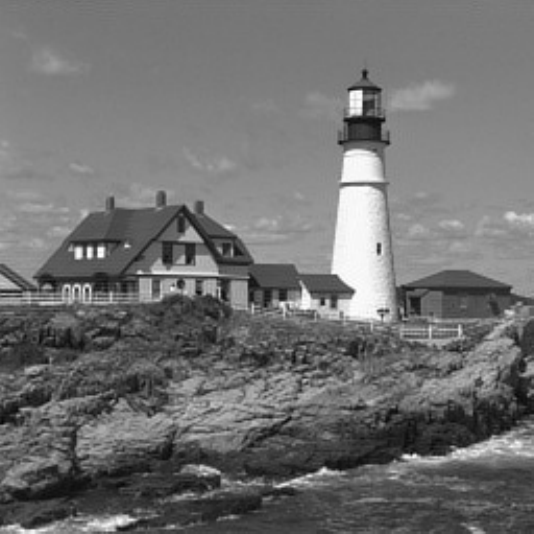} &
\includegraphics[scale=0.45]{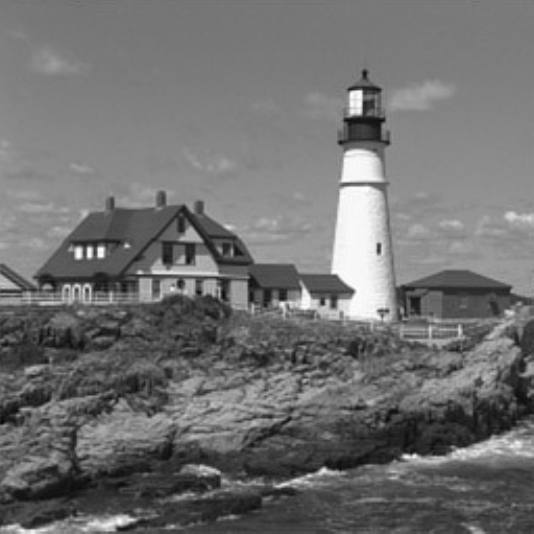} &
\includegraphics[scale=0.45]{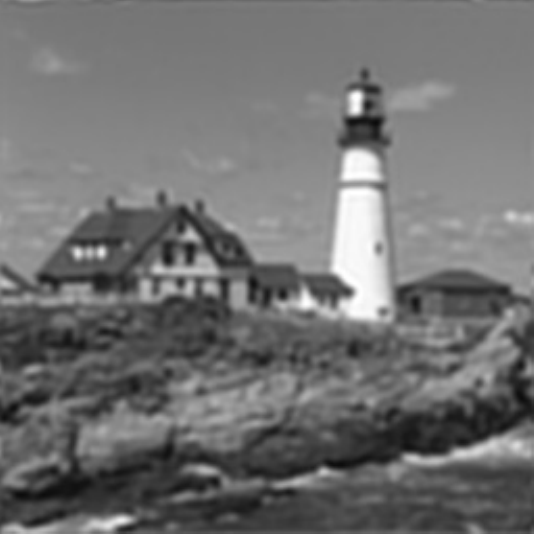} 
\end{tabular}
\end{center}
\caption{Blurred and nonblind Fried deconvolved images (first column: $C_n^2=7\times 10^{-14}m^{-2/3}$, second column: $C_n^2=2\times 10^{-13}m^{-2/3}$, third column: $C_n^2=5\times 10^{-13}m^{-2/3}$).}
\label{fig:NBDeBlur}
\end{figure}

Fig.~\ref{fig:NBDeBlur} presents the results given by the nonblind Fried algorithm on simulated blur. The different parameters are set to $D=0.5$m, $L=500$m, $\lambda=700$nm, $N=16$ and each row represents the cases $C_n^2=7\times 10^{-14}m^{-2/3}, C_n^2=2\times 10^{-13}m^{-2/3}, C_n^2=5\times 10^{-13}m^{-2/3}$. We recall that the original image can be seen in the left of Fig.~\ref{fig:ExBlur}. For each case, we build the Fried kernel $M_F$, then we blur the original image with kernel to get the simulated blurred image $f$, and finally we provide $f$ and $M_F$ to the nonblind Fried deconvolution. The first observation is that outputs of the nonblind Fried deconvolution are very close to the original image. Some details (high frequencies) are not recovered in the strong turbulence case because the blur is too important and definitively destroy these high frequencies.

\begin{figure}[!t]
\begin{center}
\begin{tabular}{cc}
\includegraphics[scale=0.6]{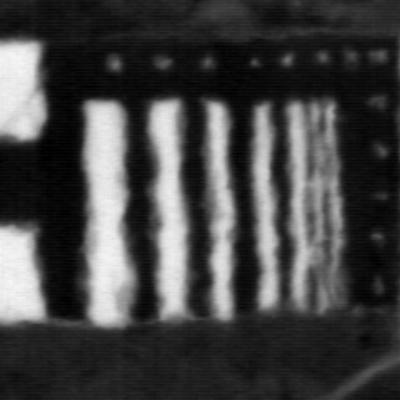} &
\includegraphics[scale=0.6]{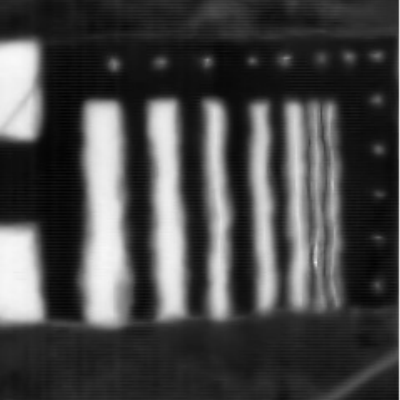} \\
\includegraphics[scale=0.48]{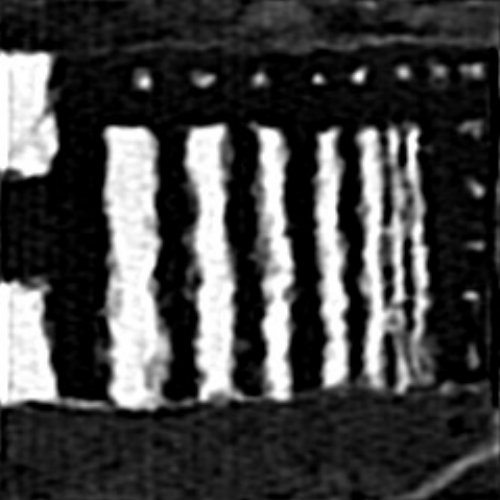} &
\includegraphics[scale=0.48]{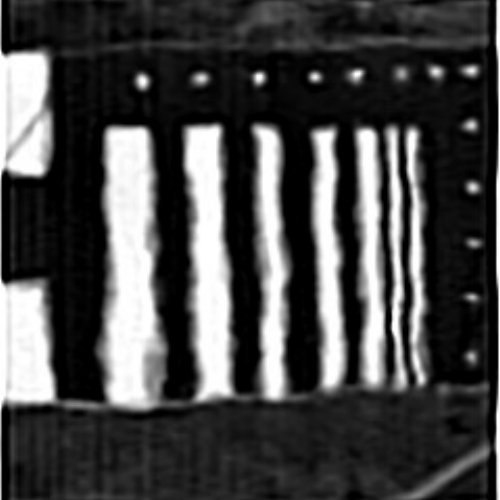}
\end{tabular}
\end{center}
\caption{Nonblind Fried deconvolution on real barchart. Original images are on top, deconvolved ones on bottom. Original acquired image is used on first column and a geometric corrected one on the second column (see text for more explanations).}
\label{fig:NBDeBlur2}
\end{figure}

\begin{figure}[!t]
\begin{center}
\begin{tabular}{cc}
\includegraphics[scale=0.6]{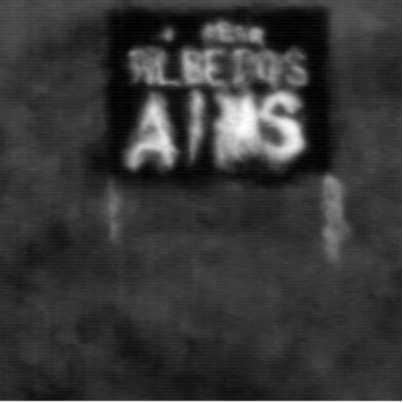} &
\includegraphics[scale=0.6]{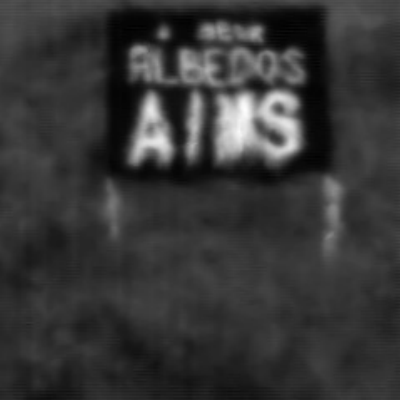} \\
\includegraphics[scale=0.48]{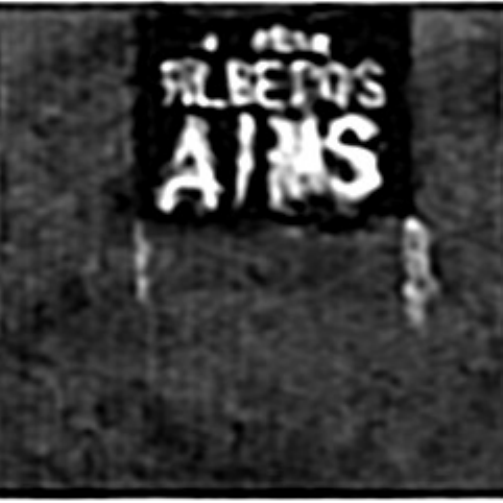} &
\includegraphics[scale=0.48]{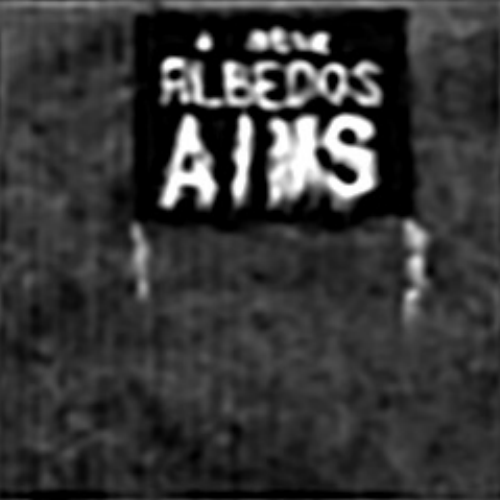}
\end{tabular}
\end{center}
\caption{Nonblind Fried deconvolution on real letter board. Original images are on top, deconvolved ones on bottom. Original acquired image is used on first column and a geometric corrected one on the second column  (see text for more explanations).}
\label{fig:NBDeBlur3}
\end{figure}

In Fig.~\ref{fig:NBDeBlur2},~\ref{fig:NBDeBlur3}, we test the nonblind Fried deconvolution on real images acquired during ground test field experiments made for turbulence characterization by the NATO group TG40 in 2005. The interesting fact is that the $C_n^2$ coefficient were acquired during each image acquisition. Here, the measured $C_n^2$ values are $1.51\times 10^{-13}m^{-2/3}$ for Fig.~\ref{fig:NBDeBlur2} and $1.91\times 10^{-13}m^{-2/3}$ for Fig.~\ref{fig:NBDeBlur3}, the observed panels were at a distance $L=1$km, the pupil diameter of the system was $D=0.05$m and the system works in the visible spectra ($\lambda\approx 700$nm). The framelet parameters are set to $\mu=1000,\eta=10$ and only two iterations were performed (experiments show that this choice gives the best visual improvements). The top row shows original images while the bottom one shows deconvolved images $\hat{u}$. The first column corresponds to the case of working with an original image directly from the output of the imaging system while the second column works with the output of the algorithm designed to reduce the turbulence geometric distortions \cite{Mao2011}. We can see great improvements particularly in edges sharpness and readibility of the letter board.

\subsection{Influence of parameters}
\begin{figure}[!t]
\begin{center}
\begin{tabular}{cc}
\includegraphics[scale=0.45]{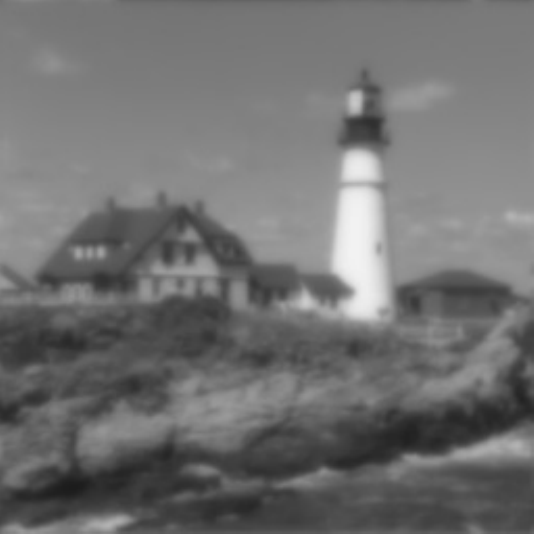} &
\includegraphics[scale=0.45]{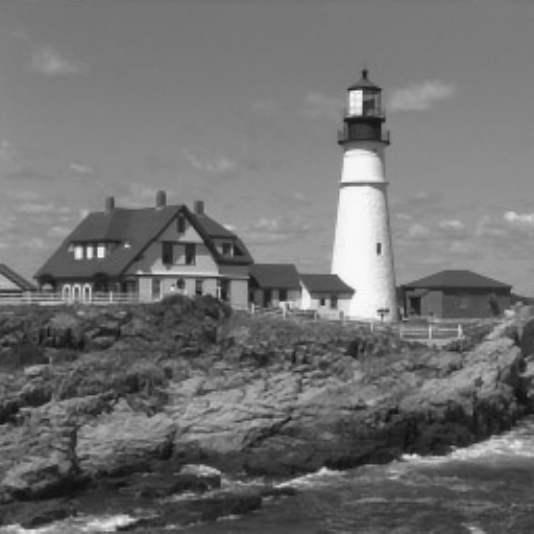} \\
\includegraphics[scale=0.45]{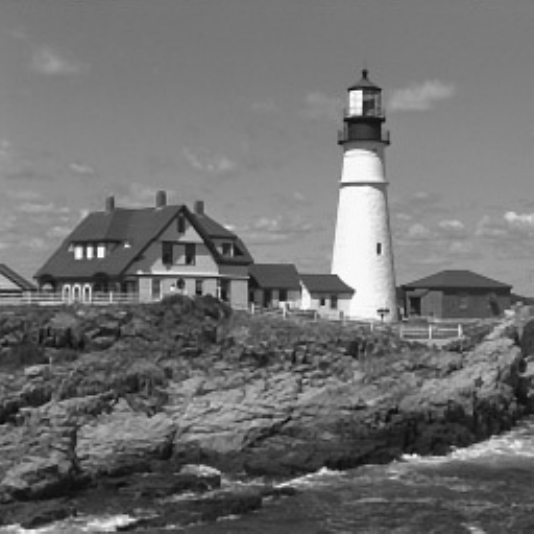} &
\includegraphics[scale=0.45]{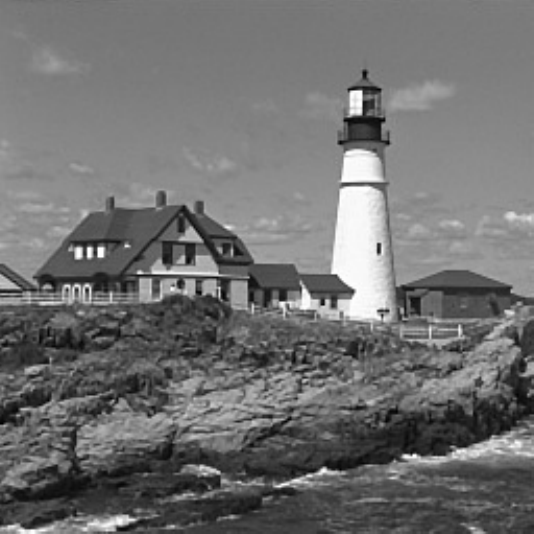}
\end{tabular}
\end{center}
\caption{Influence of the $C_n^2$ parameter on the deconvolved images. The blurred input image is on top-left, top-right is $\hat{u}$ for $C_n^2=6\times 10^{-14}m^{-2/3}$, bottom-left is $\hat{u}$ for $C_n^2=7\times 10^{-14}m^{-2/3}$ and bottom-right is $\hat{u}$ for $C_n^2=8\times 10^{-14}m^{-2/3}$.}
\label{fig:ParamCn2}
\end{figure}

\begin{figure}[!t]
\begin{center}
\begin{tabular}{cc}
\includegraphics[scale=0.45]{ParamIu} &
\includegraphics[scale=0.45]{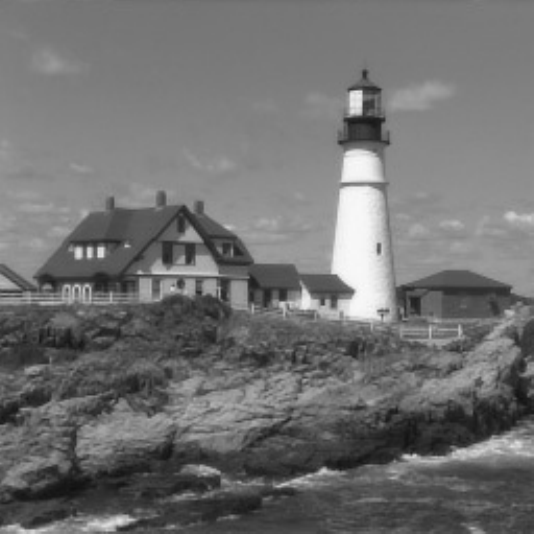} \\
\includegraphics[scale=0.45]{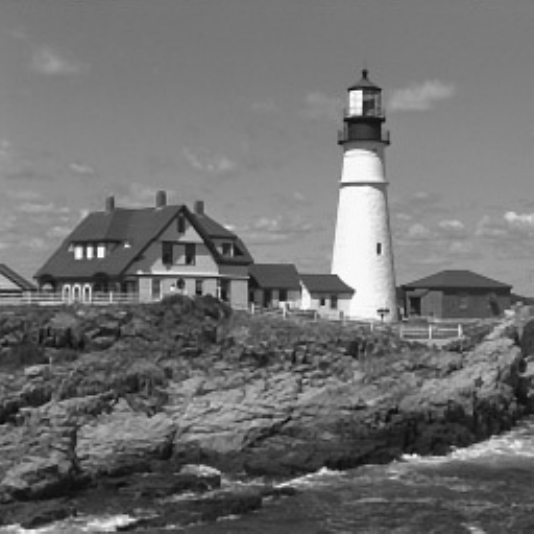} &
\includegraphics[scale=0.45]{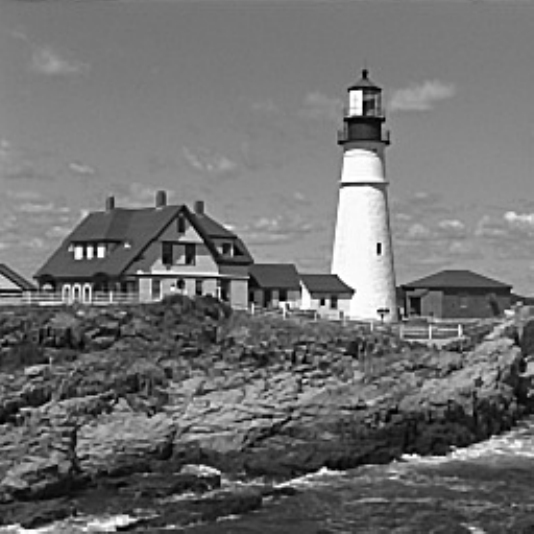}
\end{tabular}
\end{center}
\caption{Influence of the $D$ parameter on the deconvolved images. The blurred input image is on top-left, top-right is $\hat{u}$ for $D=0.04$m, bottom-left is $\hat{u}$ for $D=0.05$m and bottom-right is $\hat{u}$ for $D=0.06$m.}
\label{fig:ParamD}
\end{figure}

\begin{figure}[!t]
\begin{center}
\begin{tabular}{ccc}
\includegraphics[scale=0.45]{ParamIu} &
\includegraphics[scale=0.45]{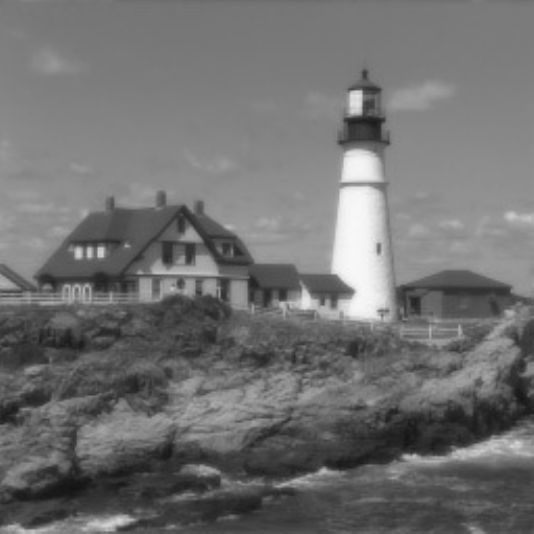} &
\includegraphics[scale=0.45]{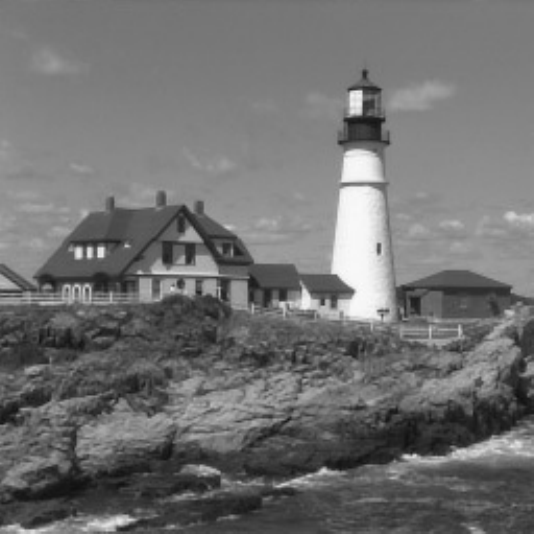} \\
\includegraphics[scale=0.45]{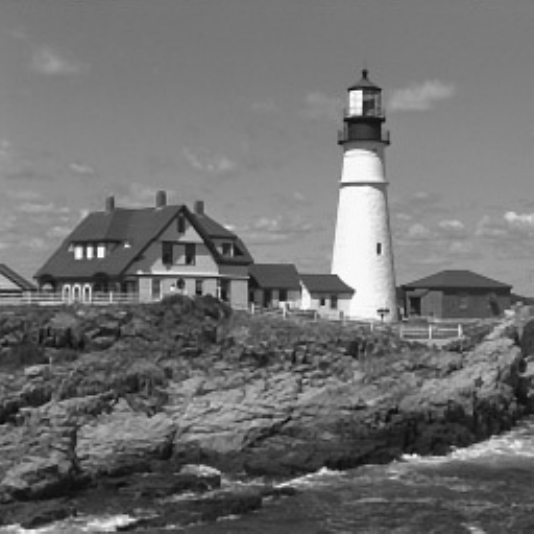} &
\includegraphics[scale=0.45]{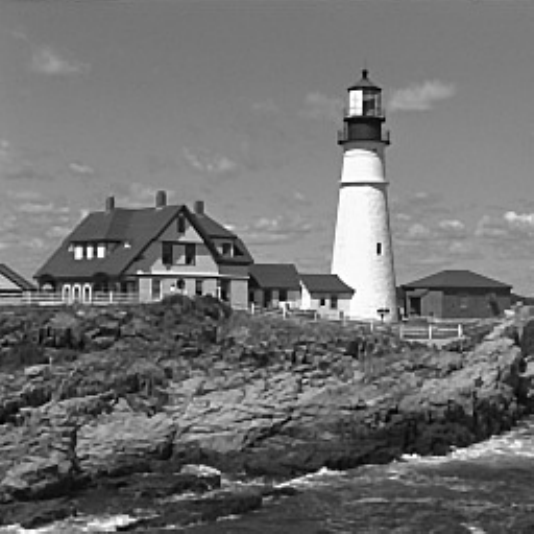} &
\includegraphics[scale=0.45]{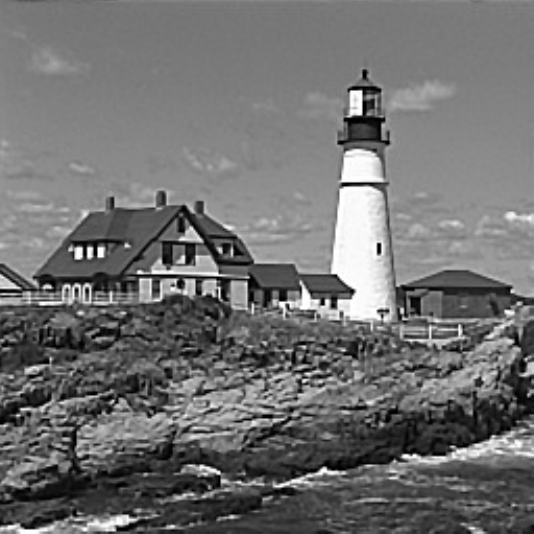}
\end{tabular}
\end{center}
\caption{Influence of the $L$ parameter on the deconvolved images. The blurred input image is on top-left, top-middle is $\hat{u}$ for $L=500$m, top-right is $\hat{u}$ for $L=800$m, bottom-left is $\hat{u}$ for $L=1000$m, bottom-middle is $\hat{u}$ for $L=1200$m and bottom-right is $\hat{u}$ for $L=1500$m.}
\label{fig:ParamL}
\end{figure}

In this section, we are interested to learn about how the deconvolved images are impacted when some errors are made on the parameter values used to build the framelet deconvolution input Fried kernel. Let's consider a visible ($\lambda=700$nm) blurred image build with the following parameters: $D=0.05$m, $L=1000$m and $C_n^2=7\times 10^{-14}m^{-2/3}$.\\
Fig.~\ref{fig:ParamCn2} shows the output images for $C_n^2=6\times 10^{-14}m^{-2/3}$, $C_n^2=7\times 10^{-14}m^{-2/3}$ (the real value) and $C_n^2=8\times 10^{-14}m^{-2/3}$ ($D$ and $L$ remain fixed). Visually, no big differences are noticeable and all outputs remain acceptable.\\
In Fig.~\ref{fig:ParamD}, we test the influence of $D$ for the values $D=0.04$m, $D=0.05$m and $D=0.06$m. Only details seem to be affected but the overall quality is excellent.\\
The last test concerns the $L$ parameter and is shown on Fig.~\ref{fig:ParamL}. We can observe under-deblurring or over-regularization in the case of large errors.\\
In practice, these tests give us promising news because it appears that even if the parameters are unknown, coarse approximations of each of them are sufficient to get an improved deconvolved image. 

\section{Blind Fried deconvolution}\label{sec:BFD}
If it seems reasonable to assume that parameters $D$,$L$ and $\lambda$ could be known in practice, the $C_n^2$ parameter is more difficult to handle as it depends on the atmospheric behavior and is not constant. In this section, we will consider the case where $C_n^2$ is unknown and needs to be estimated from the acquired blurred image $f$.
\subsection{Selection criteria}
\begin{figure}[!t]
\centering\includegraphics[scale=0.5]{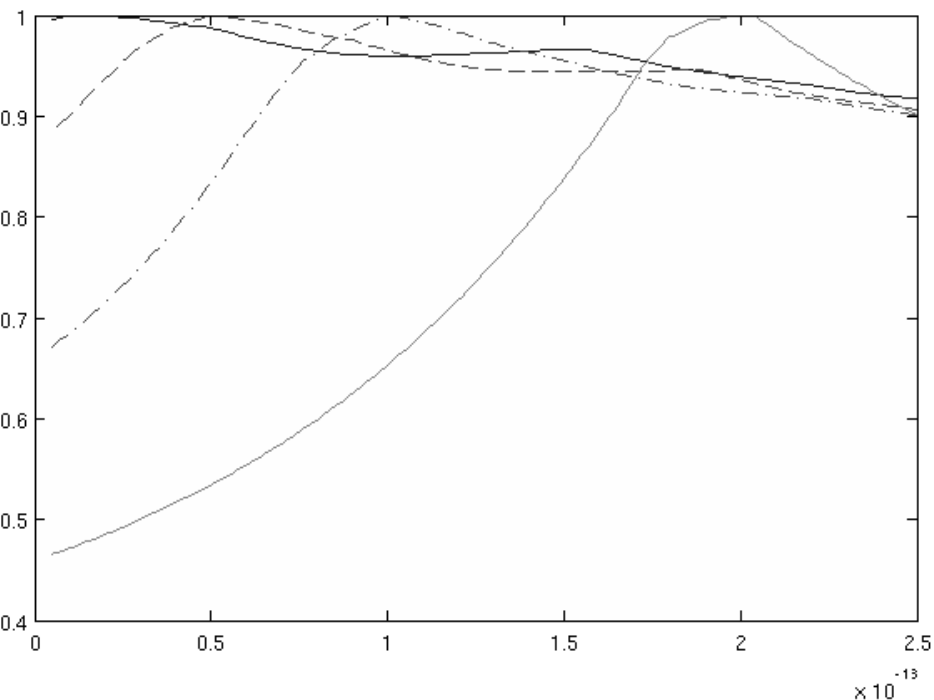} 
\caption{Curves $TV(C_n^2)$ corresponding to different blurry cases.}
\label{fig:TVnorm}
\end{figure}

The first step is to find a criteria to select the best $C_n^2$ value. From nonblind Fried deconvolution experiments we can depict the behavior of the restored image with respect to the choice of $C_n^2$ (all other parameters are assumed to be fixed). If $C_n^2$ is under-estimated, the restored image $\hat{u}$ remains blurred. If it is over-estimated, $\hat{u}$ is too much regularized and many details are removed and edges are over sharpened. This behavior naturally recalls the total variation ($TV$) defined by $TV_u=\int_{\Omega}|\nabla u|$. Indeed in the case of a blurred image, $TV$ is weak because the image gradients are over-smoothed while in the case of an image which is over-regularized (even if the gradients have more amplitude) the total amount of gradients decreases. Then it seems natural to expect that $TV$ must be maximum (with respect to the choice of $C_n^2$) for a correct restoration. This assumption can be easily verified by the following experiments: we simulate a set of different blurred images by choosing some $C_n^2$ values. Then for each case, we compute the set of deconvolved images $\hat{u}(C_n^2)$ for every $C_n^2$ choosen in an appropriate range linearly sampled which covers the typical physical range values. Then we can compute the normalized curves $TV_{\hat{u}}(C_n^2)$ for each case. Fig.~\ref{fig:TVnorm} shows the corresponding curves for cases $\hat{C}_n^2=\{1\times 10^{-14};5\times 10^{-14};10\times 10^{-14};20\times 10^{-14}\}$ $m^{-2/3}$. As expected, the position of $TV(C_n^2)$'s maximum corresponds to a good estimate of the real $C_n^2$. \\

\begin{figure}[!t]
\centering\includegraphics[scale=0.5]{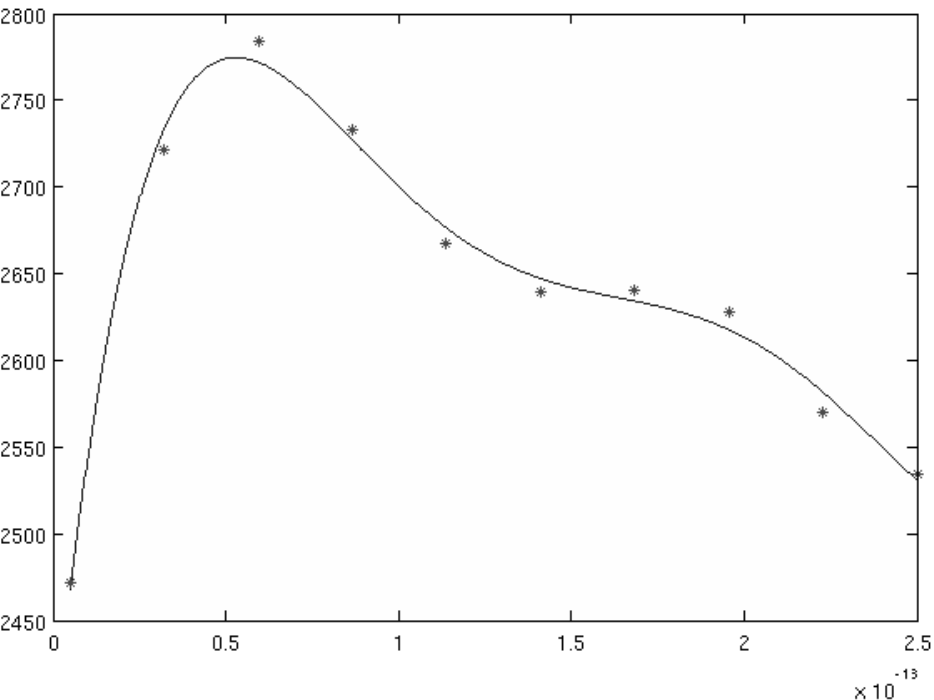} 
\caption{Polynomial approximation of the $TV(C_n^2)$ curve.}
\label{fig:polynom}
\end{figure}

In practice, it is too expensive to compute the whole $TV(C_n^2)$ curve as it needs one deconvolution per $C_n^2$ value. In order to speed up to computation to find a good estimate of $C_n^2$, we propose to compute a limited $N_{C_n^2}$ equidistant number of points of this curve. Next we find, by least square minimization, a polynomial approximation of the complete curve and finally deduce the $C_n^2$ corresponding to its maximum. Fig.~\ref{fig:polynom} shows an example of an approximation of the $TV(C_n^2)$ curve with a fifth-order polynomial from ten measured points. The corresponding $C_n^2$ estimate is $5.3.10^{-14}m^{-2/3}$ compared to the real value which was set to $5.10^{-14}m^{-2/3}$ for the experiment.

\subsection{Blind Fried deconvolution}
Now, we can depict the whole blind Fried deconvolution algorithm: given a blurry image $f$, we first compute the $N_{C_n^2}$ points of the $TV(C_n^2)$ curve which from its polynomial approximation provides an estimation $\hat{C}_n^2$ of the real $C_n^2$ coefficient. Next, we build the corresponding Fried kernel and use the framelet based deconvolution algorithm to get the final deconvolved image. This procedure is resumed on algorithm \ref{algo:bfried}

\begin{algorithm}
\caption{Blind Fried deconvolution}
\label{algo:bfried}
\begin{algorithmic}
\STATE - $D,L,\lambda$ are known. Fix the regularization parameters $\mu,\eta$.
\STATE - Compute the prescribed equidistant $N_{C_n^2}$ points of the $TV(C_n^2)$ curve for a range $C_n^2\in[C_{n,min}^2,C_{n,max}^2]$.
\STATE - Find the polynomial approximation by least square minimization.
\STATE - Estimate $\hat{C}_n^2$ from the maximum of the polynomial approximation.
\STATE - Build the Fried kernel $M_F(\omega)$ with the known parameters and $\hat{C}_n^2$.
\STATE - Use the framelet nonblind deconvolution algorithm with $M_F(\omega)$ to find the final restored image $\hat{u}$.
\end{algorithmic}
\end{algorithm}

\subsection{Experiments}
\begin{figure}[!t]
\begin{center}
\begin{tabular}{ccc}
\includegraphics[scale=0.45]{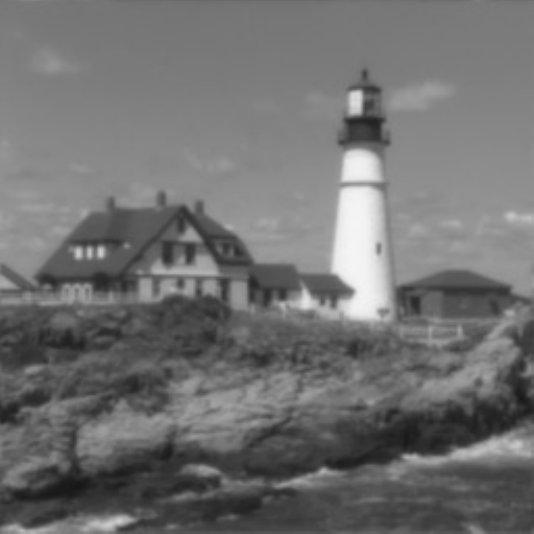} &
\includegraphics[scale=0.45]{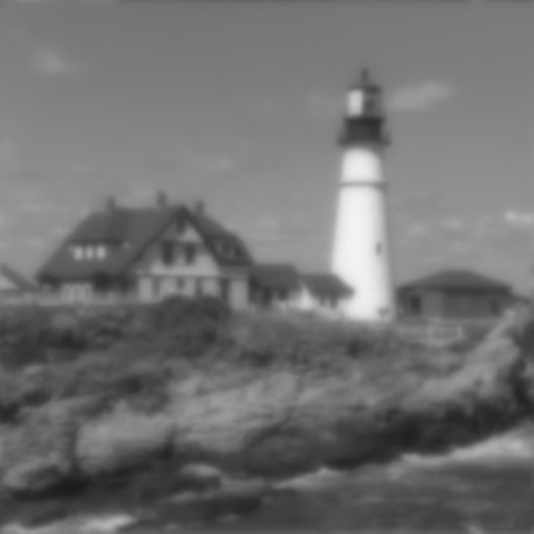} &
\includegraphics[scale=0.45]{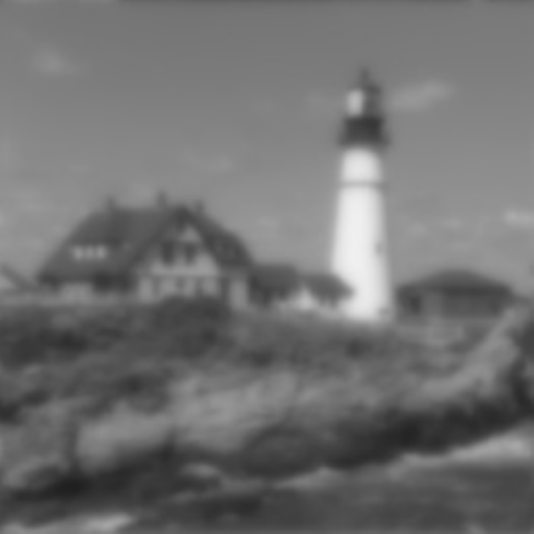} \\
\includegraphics[scale=0.45]{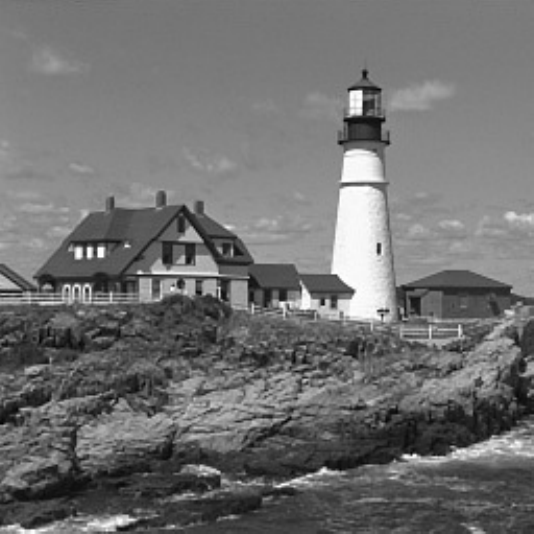} &
\includegraphics[scale=0.45]{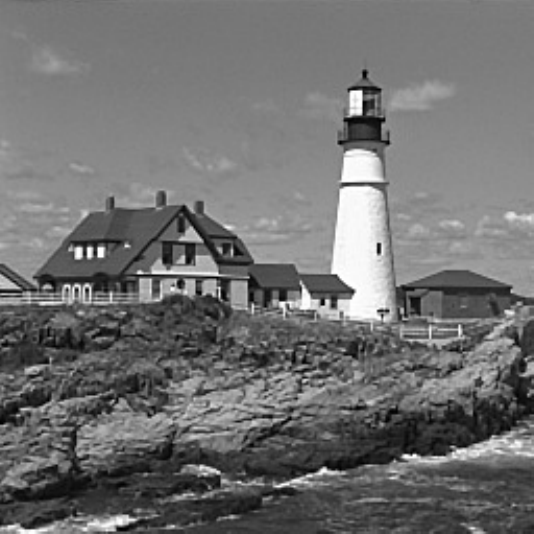} &
\includegraphics[scale=0.45]{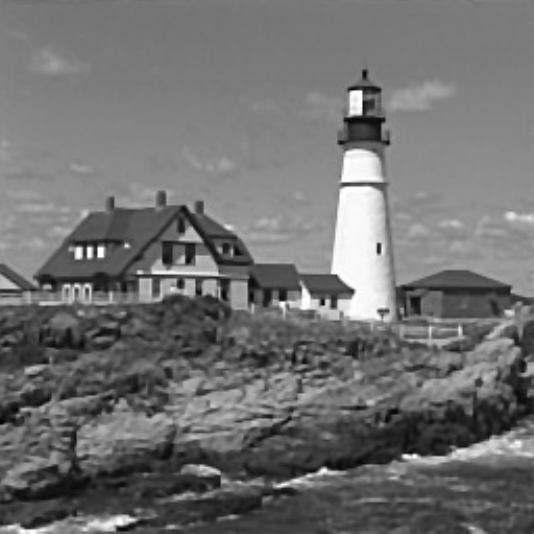} 
\end{tabular}
\end{center}
\caption{Blurred and blind Fried deconvolved images (first column: $C_n^2=1.5\times 10^{-14}m^{-2/3}$, second column: $C_n^2=7.6\times 10^{-13}m^{-2/3}$, third column: $C_n^2=13.8\times 10^{-13}m^{-2/3}$).}
\label{fig:BDeBlur}
\end{figure}

In this section, we present different experiments based on the blind Fried deconvolution described previously. In all experiments, the different parameters are set to $C_{n,min}^2=0.5.10^{-14}m^{-2/3}, C_{n,max}^2=2.5.10^{-13}m^{-2/3}$, $N_{iter}=3,N_{C_n^2}=10$ and $\eta=10$ respectively. In this paper, we also consider grayscale visible images only and consequently we set $\lambda=700$nm. The first experiment is based on simulated blur. We choose $L=500$m, $D=0.07$m, $\mu=10^5$ and a set of $C_n^2\in\{1.5\times 10^{-14}m^{-2/3};7.6\times 10^{-14}m^{-2/3};13.8\times 10^{-14}m^{-2/3}\}$. Table~\ref{tab:cn2} gives a comparison between the real $C_n^2$ values used in the simulation and their corresponding estimates from the polynomial approximation $TV(C_n^2)$ curve. Even if we can see some deviations in these estimations, Fig.~\ref{fig:BDeBlur} shows that the deconvolved images are significantly improved compared to the blurred versions.

\begin{table}
\begin{center}
\begin{tabular}{|c|c|c|c|}\hline
Real $C_n^2$ & $1.5\times 10^{-14}$ & $7.6\times 10^{-14}$ & $13.8\times 10^{-14}$ \\ \hline
Estimated $\hat{C}_n^2$ & $2.1\times 10^{-14}$ & $9.71\times 10^{-14}$ & $15.5\times 10^{-14}$ \\ \hline
\end{tabular}
\end{center}
\caption{Real and estimated $C_n^2$ values (in $m^{-2/3}$) computed from the simulated images.}
\label{tab:cn2}
\end{table}

\begin{figure}[!t]
\begin{center}
\begin{tabular}{cc}
\includegraphics[scale=0.6]{F51_14112005_0800} &
\includegraphics[scale=0.6]{UnwrapMire_14112005_0800} \\
\includegraphics[scale=0.48]{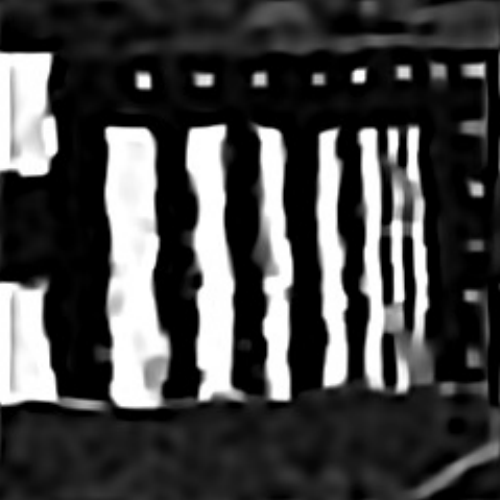} &
\includegraphics[scale=0.48]{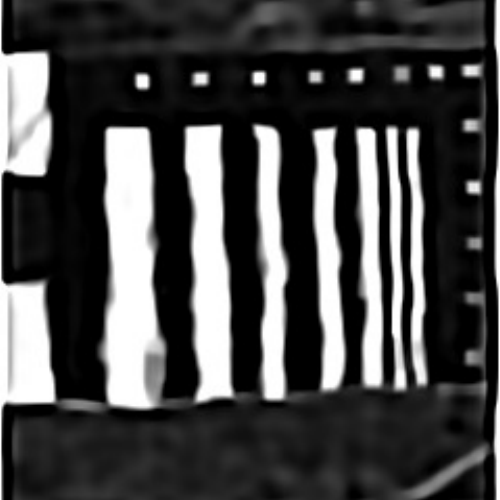}
\end{tabular}
\end{center}
\caption{Blind Fried deconvolution on real barchart. Original images are on top, deconvolved ones on bottom. Original acquired image is used on first column and a geometric corrected one on the second column (see text for more explanations).}
\label{fig:BDeBlur2}
\end{figure}

\begin{figure}[!t]
\begin{center}
\begin{tabular}{cc}
\includegraphics[scale=0.6]{F21_14112005_0804} &
\includegraphics[scale=0.6]{Unwrap_14112005_0804} \\
\includegraphics[scale=0.48]{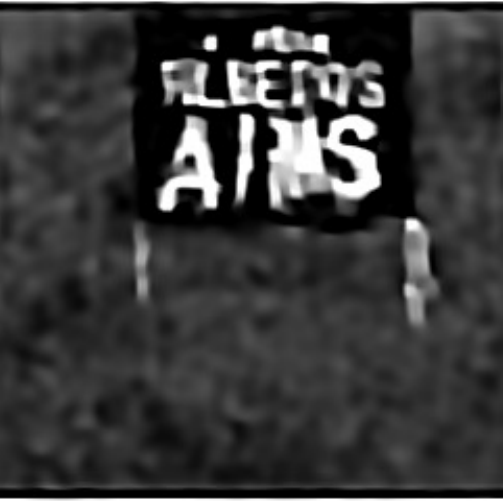} &
\includegraphics[scale=0.48]{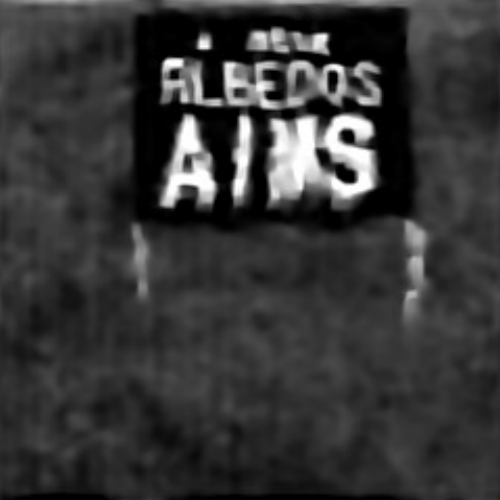}
\end{tabular}
\end{center}
\caption{Blind Fried deconvolution on real letter board. Original images are on top, deconvolved ones on bottom. Original acquired image is used on first column and a geometric corrected one on the second column (see text for more explanations).}
\label{fig:BDeBlur3}
\end{figure}

The second experiment is based on the real images used in Fig.~\ref{fig:NBDeBlur2} and \ref{fig:NBDeBlur3}. We keep the same parameters, except for $\mu$ which is set to $10^3$. Table~\ref{tab:rcn2} gives the real $C_n^2$ coefficients and their estimations, Fig.~\ref{fig:BDeBlur2} and \ref{fig:BDeBlur3} show the original images and their deconvolved versions. We can see that the final output has great improvements, especially when the deconvolution is applied to the geometrically corrected images. This seems to confirm that a combination of a geometric distortion correction plus a specific deblurring is a good combination to deal with the turbulence problem. We notice for the barchart image that the $C_n^2$ coefficient is in both case estimated to its maximum value ($C_n^2=C_{n,max}^2$). But this is not surprising because the image does not have any details like textures and the phenomena of over-regularization does not make sense in this case. On the letter board case, we can see in Table~\ref{tab:rcn2}, that the $C_n^2$ estimation is lower for the geometric corrected image. This is because the geometric correction algorithm already gives some improved and sharpened images and then virtually reduce the blur.

\begin{table}[h]
\caption{Real and estimated $C_n^2$ values (in $m^{-2/3}$) computed for real images.}
\label{tab:rcn2}
\begin{center}
\begin{tabular}{|c|c|c|}\hline
& Original1 & Unwarped1 \\ \hline 
Real $C_n^2$ & $15.1\times 10^{-14}$ &  \\ \hline
Estimated $\hat{C}_n^2$ & $25\times 10^{-14}$ & $25\times 10^{-14}$ \\ \hline
& Original2 & Unwarped2 \\ \hline
Real $C_n^2$ & $19.1\times 10^{-14}$ & \\ \hline
Estimated $\hat{C}_n^2$  & $20.7\times 10^{-14}$ & $11.6\times 10^{-14}$ \\ \hline
\end{tabular}
\end{center}
\end{table}

\section{Conclusion - Future work}
In this paper, we introduce the possibility to use the analytical formulation of the Fried kernel which model the atmosphere effects on images to do some deconvolution and improve the quality of long range images. We propose both nonblind and blind methods with respect to a difficult parameter to deal with: the refractive index structure $C_n^2$ which represent the level of turbulence of the atmosphere. Finally, we get very simple and efficient algorithms which are easy to implement.\\

In a future work, we will investigate a fusion of both geometric distortion correction and Fried deconvolution in a unified framework to solve the problem of atmospheric turbulence.\\

\acknowledgements
The authors wants to thank Dr. Richard Espinola from the U.S. Army RDECOM CERDEC NVESD for letting us know about the existence of the Fried kernel. We thanks the other members of the NATO SET156 (ex-SET072 RTG40) Group for the opportunity of using the data collected during the 2005 New Mexico's field trials. This work is supported by the following grants: NSF DMS-0914856, ONR N00014-08-1-119, ONR N00014-09-1-360.

\end{document}